\documentclass[10pt,twocolumn]{proc}
\usepackage{lipsum}
\usepackage{bbm}
\usepackage{algorithm}
\usepackage{algpseudocode}
\usepackage[english]{babel}
\usepackage{amsmath}
\usepackage{graphicx}
\usepackage{fancyhdr}
\usepackage[margin=0.5in]{geometry}

\newcommand{\argmax}{\operatornamewithlimits{argmax}}
\providecommand{\keywords}[1]{\textbf{\textit{Index terms---}} #1}

\begin{document}
\title{Semi-Supervised Nonlinear Distance Metric Learning via Forests of Max-Margin Cluster Hierarchies}

\author{
David M. Johnson \\
davidjoh@buffalo.edu \and
Caiming Xiong \\
cxiong@buffalo.edu \and
Jason J. Corso\\
jcorso@buffalo.edu
\thanks{Authors are with the Department of Computer Science and Engineering, SUNY at Buffalo, Buffalo, NY, 14260-2500}
}

\pagestyle{fancy}
\fancyhead{}
\fancyfoot{}

\fancypagestyle{firststyle}
{
   \fancyhf{}
   \fancyfoot{}
   \fancyhead[HL]{\footnotesize{MANUSCRIPT SUBMITTED TO SIGKDD ON 21 FEB 2014}}
}

\renewcommand{\headrulewidth}{0pt}
\renewcommand{\footrulewidth}{0pt}

\fancyhead[HL]{\footnotesize{MANUSCRIPT SUBMITTED TO SIGKDD ON 21 FEB 2014}}
\fancyhead[HR]{\thepage}


\maketitle
\begin{abstract}
Metric learning is a key problem for many data mining and machine learning applications, and has long been dominated by Mahalanobis methods. Recent advances in nonlinear metric learning have demonstrated the potential power of non-Mahalanobis distance functions, particularly tree-based functions.  We propose a novel nonlinear metric learning method that uses an iterative, hierarchical variant of semi-supervised max-margin clustering to construct a forest of cluster hierarchies, where each individual hierarchy can be interpreted as a weak metric over the data.  By introducing randomness during hierarchy training and combining the output of many of the resulting semi-random weak hierarchy metrics, we can obtain a powerful and robust nonlinear metric model. This method has two primary contributions: first, it is semi-supervised, incorporating information from both constrained and unconstrained points.  Second, we take a relaxed approach to constraint satisfaction, allowing the method to satisfy different subsets of the constraints at different levels of the hierarchy rather than attempting to simultaneously satisfy all of them.  This leads to a more robust learning algorithm.  We compare our method to a number of state-of-the-art benchmarks on $k$-nearest neighbor classification, large-scale image retrieval and semi-supervised clustering problems, and find that our algorithm yields results comparable or superior to the state-of-the-art, and is significantly more robust to noise.
\end{abstract}

\keywords{Metric learning, nonlinear, semi-supervised, max-margin clustering, hierarchical clustering}

\section{Introduction}
\thispagestyle{firststyle}
Many elemental data mining problems---nearest neighbor classification, retrieval, clustering---are at their core dependent on the availability of an effective means of measuring pairwise distance.  Ad hoc selection of a metric, whether by relying on a standard such as Euclidean distance or attempting to select a domain-appropriate kernel, is unreliable and inflexible.  We thus approach metric selection as a learning problem, and attempt to train strong problem-specific distance measures using data and semantic information.

A wide range of methods have been proposed to address this learning problem, but the field has traditionally been dominated by algorithms that assume a linear model of distance, particularly Mahalanobis metrics~\cite{bellet2013survey}.  Linear methods have primarily benefited from two advantages.  First, they are generally easier to optimize, allowing for faster learning and in many cases a globally optimal solution to the proposed problem~\cite{davis2007information, Shen2009PSD, blitzer2005distance, ying2012distance}.  Second, they allow the original data to be easily projected into the new metric space, meaning the metric can be used in conjunction with other methods that operate only on an explicit feature representation (most notably approximate nearest neighbor methods---needed if the metric is to be applied efficiently to large-scale problems).

However, for many types of data a linear approach is not appropriate.  Images, videos, documents and histogram representations of all kinds are ill-suited to linear models.  Even an ideal Mahalanobis metric will be unable to capture the true semantic structure of these types of data, particularly over larger distances where local linearity breaks down.  Kernelized versions of popular Mahalanobis methods~\cite{davis2007information, chatpatanasiri2010new} have been proposed to handle such data, but these approaches have been limited by high complexity costs.  For this reason, researchers have begun to seek alternate metric models that are inherently capable of handling nonlinear data.  

These nonlinear metrics are necessarily a broad class of models, encompassing a range of learning modalities and metric structures.  One early example of nonlinear metrics (for facial recognition, in this case) by Chopra et al.~\cite{chopra2005learning} was based on deep learning strategies.  The method was effective, but required long training times and extensive tuning of hyperparameters.  Other methods sought to resolve the problem by taking advantage of local linearity in the data, and learning multiple localized linear metrics \cite{frome2006image, frome2007learning, weinberger2008fast, zhan2009learning}.  These techniques have generally proven superior to single-metric methods, but have also tended to be expensive.  

Most recently, several works have explored metrics that take advantage of tree structures to produce flexible nonlinear transformations of the data.  Kedem et al.~\cite{kedem2012non} proposed a method that trained a set of gradient-boosted regression trees and added the regression outputs of each region directly to the data, producing an explicit nonlinear transformation that shifted similar points together and dissimilar points apart.  However, this method relies on performing regression against a $d$-dimensional gradient vector, and may thus be prone to overfitting when $d$ is large relative to the number of training samples.

Finally, our previous work in this area~\cite{xiong2012random} formulated the pairwise-constrained metric learning problem as a pair-classification problem, and solved it by direct application of random forests, yielding an implicit nonlinear transformation of the data.  However, while this metric could be trained efficiently, it suffered from poor scalability at inference time due to the lack of an explicit feature representation, which made common metric tasks such as nearest neighbor search expensive on larger datasets.

In order to overcome the limitations of these methods, we propose a novel tree-based nonlinear metric with several advantages over existing algorithms.  Our metric first constructs a model of the data by computing a forest of semi-random cluster hierarchies, where each tree is generated by iteratively applying a partially-randomized binary semi-supervised max-margin clustering objective.  As a result, each tree directly encodes a particular model of the data's full semantic structure, and the structure of the tree itself can thus be interpreted as a weak metric.  By merging the output from a forest of these weak metrics, we can produce a final metric model that is powerful, flexible, and resistant to overtraining (due to the independent and semi-random nature of the hierarchy construction).  

This methodology provides two significant contributions:  first, unlike previous tree-based nonlinear metrics, it is semi-supervised, and can incorporate information from both constrained and unconstrained points into the learning algorithm.  This is an important advantage in many problem settings, particularly when scaling to larger datasets where only a tiny proportion of the full pairwise constraint set can realistically be collected or used in training.  

Second, the iterative, hierarchical nature of our training process allows us to relax the constraint satisfaction problem.  Rather than attempting to satisfy every available constraint simultaneously, at each hierarchy node we can optimize an appropriate constraint subset to focus on, leaving others to be addressed lower in the tree (or in other hierarchies in the forest).  By selecting constraints in this way, we can avoid situations where we are attempting to satisfy incoherent constraints \cite{wagstaff2006constrained}, and thereby better model hierarchical data structures.  We can also obtain an algorithm that is more robust to noisy data (see experiments in Section \ref{sec:exp_class}).

Additionally, we propose a scalable and highly accurate algorithm for obtaining approximate nearest neighbors within our learned metric's space.  This renders the metric tractable for large-scale retrieval or nearest-neighbor classification problems, and overcomes a major limitation our previous tree-based metric.

The remainder of this paper is organized as follows: in Section~\ref{sec:HFDover}, we describe in detail our formulation for hierarchy-forest-based metric learning.  In Section~\ref{sec:mmc} we discuss max-margin clustering and describe our formulation of semi-super\-vised max-margin clustering for hierarchy learning.  In Section~\ref{sec:hfdnn} we describe our method for fast approximate in-metric nearest nearest-neighbor retrieval.  In Section~\ref{sec:complex} we provide a complexity analysis of our method, and in Section~\ref{sec:exp} we show experimental results in which our method compares favorably to the state-of-the-art in metric learning.


\section{Semi-supervised max-margin hierarchy forests}
\label{sec:HFDover}

In this section we describe in detail our Hierarchy Forest Distance (HFD) model, as well as our procedures for training and inference.  The structure of the HFD model draws some basic elements from random forests~\cite{breiman2001random}, in that it is composed of $T$ trees trained independently in a semi-random fashion, with individual nodes in the trees defined by a splitting function that divides the local space into two or more segments.  Each hierarchy tree represents a distance function $\mathcal{H}(a,b)$, and the overall distance function is
\begin{equation}
\label{eq:forest}
D(a,b) = \frac{1}{T}\sum_{t=1}^T \mathcal{H}_t(a,b) \enspace.
\end{equation}
However, HFD is conceptually distinct from random forests (and the Random Forest Distance (RFD) metric~\cite{xiong2012random}) in that the individual components of the forest represent cluster hierarchies rather than decision trees.  We discuss this distinction and its implications in Section~\ref{sec:HFDtrain}.

\subsection{Hierarchy forest distance}
\label{sec:hfd}

The full hierarchy forest distance is effectively the mean of a number of weak distance functions $\mathcal{H}_t$, each corresponding to one hierarchy in the forest.  These distance functions, in turn, are representations of the structure of the individual hierarchies---the further apart two instances fall within a hierarchy, the greater the distance between them.  Specifically, we formulate each metric as a modified form of the hierarchy distance function we previously proposed for use in hierarchy comparison~\cite{johnson2013comprehensive}:
\begin{equation}
\label{eq:metric}
\mathcal{H}_t(a,b) =
\begin{cases}
0 & \text{if } \mathcal{H}_{tl(a,b)} \text{is a leaf node}\\
p_t(a,b) \cdot \frac{\left| \mathcal{H}_{tl(a,b)}\right|}{N} & \text{otherwise,}
\end{cases}
\end{equation}
where $\mathcal{H}_t$ represents a particular hierarchy, $a$ and $b$ are input points, $\mathcal{H}_{tl}$ denotes the $l^{th}$ node in  $\mathcal{H}_t$,  $\mathcal{H}_{tl(a,b)}$ is the the smallest (i.e. lowest) node in $\mathcal{H}_t$ that contains both $a$ and $b$ and $\left| \mathcal{H}_{tl(a,b)}\right|$ represents the number of training points (out of the whole training set of size $N$) contained in that node's subtree.  Pairs that share a leaf node are given a distance of 0 because they are maximally similar under $\mathcal{H}_t$, and the minimum size of leaf nodes is defined by parameter, not by data.  Each non-leaf node $\mathcal{H}_{tl}$ is assigned (via max-margin clustering) a projection function $\mathcal{P}_{tl}$ and associated binary linear discriminant $\mathcal{S}_{tl}$  that divides the data in that node between the two child nodes. $p_t(a,b)$ is a certainty term determined by the distance of the projected points $a$ and $b$ from the decision hyperplane at $\mathcal{H}_{tl(a,b)}$:
\begin{equation}
p_t(a,b) = \frac{1}{1+\exp(\alpha \cdot \mathcal{P}_{tl(a,n)}(x_a))} - \frac{1}{1+\exp(\alpha \cdot \mathcal{P}_{tl(a,b)}(x_b))} \enspace,
\end{equation}
where $\alpha$ is a hyperparameter that controls the sensitivity of $p$.  Thus, $p$ ranges from 0 to 1, approaching 0 when the projections of both $a$ and $b$ are near the decision boundary, and 1 when both are far away.  The full distance formulation for a hierarchy is also confined to this range, with a distance approaching 1 corresponding to points that are widely separated at the root node, and 0 to points that share a leaf node.

\subsection{HFD learning and inference}
\label{sec:HFDtrain}

The fact that the trees used in HFD represent cluster hierarchies rather than decision trees has significant implications for HFD training, imposing stricter requirements on the learned splitting functions.  While the goal of decision tree learning is ultimately to yield a set of pure single-class leaf nodes, a cluster hierarchy instead seeks to accurately group data elements at \emph{every} level of the tree.  Thus, if the hierarchy learning algorithm divides the data poorly at or near the root node, there is no way for it to recover from this error later on.  This is partially mitigated by learning a forest in place of a single tree, but even in this case the \emph{majority} of hierarchies in the forest must correctly model the high-level semantic relationship between any two data elements.

For this reason, HFD requires a robust approach to the hierarchy splitting problem that reliably generates semantically meaningful splits.  Additionally, in order to allow for efficient metric inference, our splitting algorithm must generate explicit and efficiently evaluable splitting functions at each node.

Given these constraints, we approach the hierarchy learning problem as a series of increasingly fine-grained flat semi-supervised clustering problems, and we solve these flat clustering problems via max-margin clustering (MMC)~\cite{xu2004maximum, zhao2008efficient,zhang2009maximum,hoai2012maximum}.  Max-margin clustering has a number of advantages that make it ideal for our problem:
\begin{itemize}
\item max-margin and large-margin methods have been prov\-en effective in the metric learning domain~\cite{blitzer2005distance, XiJoCoECDM2012b,shen2010scalable}
\item MMC returns a simple and explicit splitting function which can be applied to points outside the initial clustering
\item MMC (including semi-supervised MMC) can be solved in linear time~\cite{hoai2012maximum, hu2008maximum,zeng2012semi}
\end{itemize}
We employ a novel relaxed form of semi-supervised MMC, which uses pairwise must-link (ML) and cannot-link (CL) constraints to improve semantic clustering performance. Constraints of this type indicate either semantic similarity (ML) or dissimilarity (CL) between pairs of points, and can be provided by themselves.

We describe our semi-supervised MMC technique in Section~\ref{sec:mmc}.

\subsubsection{Training algorithm}
\label{sec:HFDtrainalg}
We train each tree in the HFD model independently, with each tree using the same data and constraint sets. Training is hence easily parallelized.  Assume an unlabeled training dataset $\mathbf{X}^0$ and pairwise constraint set $\mathcal{L}^0$.  Denote a must-link constraint set $\mathcal{L}^0_\mathcal{M}$ and cannot-link constraint set $\mathcal{L}^0_\mathcal{C}$, such that $\mathcal{L}^0 = \mathcal{L}^0_\mathcal{M} \cup \mathcal{L}^0_\mathcal{C}$.

Training of individual trees proceeds in a top-down manner.  At each node $\mathcal{H}_{tl}$ we begin by selecting a local feature subset $\mathcal{K}_{tl}$ by uniformly sampling $d_k < d$ features from the full feature set.  We then replace each $\mathbf{x}_j \in \mathbf{X}^{tl}$ with $\begin{bmatrix} \mathbf{x}^{\mathcal{K}_{tl}}_j & 1 \end{bmatrix} \in \mathbf{X}_{\mathcal{K}}^{tl}$.

For each node $\mathcal{H}_{tl}$, our split function learning algorithm can operate in either a semi-supervised or unsupervised mode, so before we begin learning we must check for constraint availability.  We require at least 1 cannot-link constraint in order to carry out semi-supervised MMC, so we check whether $\mathcal{L}^{tl}_\mathcal{C} = \emptyset$, and then apply either semi-supervised or unsupervised MMC (see Section \ref{sec:mmc}) to $\mathbf{X}_{\mathcal{K}}^{tl}$ and $\mathcal{L}^{tl}$. The output of our split learning algorithm is the weight vector $\mathbf{w}_{tl}$, which, along with $\mathcal{K}_{tl}$, forms the splitting function $\mathcal{S}_{tl}$:
\begin{align}
\mathcal{P}_{tl}(\mathbf{x}) &= \mathbf{w}_{tl}^T \begin{bmatrix} \mathbf{x}^{\mathcal{K}_{tl}}_j & 1 \end{bmatrix}\\
\label{eq:split}
\mathcal{S}_{tl}(\mathbf{x}) &= \
\begin{cases}
\text{send $\mathbf{x}$ left} & \mathcal{P}_{tl}(\mathbf{x}) \leq 0 \\
\text{send $\mathbf{x}$ right} & \mathcal{P}_{tl}(\mathbf{x}) > 0
\end{cases} \enspace.
\end{align}

We then apply $\mathcal{S}_{tl}$ to divide $\mathbf{X}_{tl}$ among $\mathcal{H}_{tl}$'s children.  After this, we must also propagate the constraints down the tree.  We do this by iterating through $\mathcal{L}_{tl}$ and checking the point membership of each child node $\mathcal{H}_{tj}$---if $\mathbf{X}_{tj}$ contains both points covered by a constraint, then we add that constraint to $\mathcal{L}_{tj}$.  

As a result, constraints in $\mathcal{L}_{tl}$ whose constrained points are separated by $\mathcal{H}_{tl}$'s splitting function effectively disappear in the next level of the hierarchy.  This results in a steady narrowing of the constraint-satisfaction problem as we reach further down the tree, in accordance with the progressively smaller regions of the data space we are processing.  We continue this process until we reach a stopping point (in our experiments, a minimum node size threshold), falling back on unsupervised MMC as we exhaust the relevant cannot-link constraints.

\subsubsection{Inference}
\label{sec:HFDinf}

Metric inference on learned HFD structures is straightforward.  We feed two points $\mathbf{x}_1$ and $\mathbf{x}_2$ to the metric and track their progress down each tree $\mathcal{H}_t$.  At each node $\mathcal{H}_{tl}$, we compute $\mathcal{S}_{tl}(\mathbf{x}_1)$ and $\mathcal{S}_{tl}(\mathbf{x}_2)$.  If $\mathcal{S}_{tl}(\mathbf{x}_1) = \mathcal{S}_{tl}(\mathbf{x}_2)$, we continue the process in the indicated child node.  If not, then we have found $\mathcal{H}_{tl(\mathbf{x}_1, \mathbf{x}_2)}$, so we compute and return $\mathcal{H}_t(\mathbf{x}_1, \mathbf{x}_2)$ as described in (\ref{eq:metric}).  The results from each tree are then combined as per (\ref{eq:forest}).

\section{Learning splitting functions}
\label{sec:mmc}

In order to learn strong, optimized splitting functions at each hierarchy node, our method relies on the Max-Margin Clustering (MMC) framework.  In most nodes, our method uses semi-supervised MMC (SSMMC) to incorporate pairwise constraint information into the split function learning process.  Below, we describe a state-of-the art SSMMC formulation, as well as our own novel modifications to this formulation that allow it to function in our hierarchical problem setting.

\subsection{Semi-supervised max-margin clustering}
\label{sec:ssmmc}

SSMMC incorporates a set of must-link ($\mathcal{L_M}$) and cannot-link ($\mathcal{L_C}$) pairwise semantic constraints into the clustering problem.  Thus, where unsupervised MMC seeks only to maximize the cluster assignment margin of each point, SSMMC includes an additional set of margin terms reflecting the satisfaction of each pairwise constraint.

For convenience, first define the following function representing the joint projection value of two different points onto two particular cluster labels\footnote{Note that we will periodically refer to $\phi_t$ or $\phi_r$---these simply indicate that the $\phi$ function is using the temporary $\mathbf{w}$ value at the given iteration (e.g. $\mathbf{w}^{(t)}$).}: 
\begin{equation}
\phi(\mathbf{x}_1, \mathbf{x}_2, y_1, y_2) =  y_1\mathbf{w}^T\mathbf{x}_1 + y_2\mathbf{w}^T\mathbf{x}_2
\end{equation}
The semi-supervised MMC problem~\cite{zeng2012semi} is then formulated as:
\begin{equation}
\label{eq:mrmmc}
\begin{gathered}
\shoveleft \min_{\mathbf{w}, \eta, \xi} \frac{\lambda}{2}\|\mathbf{w}\|^2 + \frac{1}{L_\mathcal{M} + L_{\mathcal{C}}}\left(\sum_{j\in \mathcal{L_M}} \eta_j + \sum_{j\in \mathcal{L_C}} \eta_j \right) + \frac{C}{U}\sum_{i=1}^U\xi_i \\
\begin{aligned}
&\text{s.t.}&&  \\
\noalign{\quad $\forall j \in \mathcal{L_M}, \forall s_{j1}, s_{j2} \in \{-1, 1\}, s_{j1} \neq s_{j2} :$}
&&& \text{ } \max_{z_{j1}=z_{j2}} \phi(\mathbf{x}_{j1}, \mathbf{x}_{j2}, z_{j1}, z_{j2}) - \phi(\mathbf{x}_{j1}, \mathbf{x}_{j2}, s_{j1}, s_{j2})\\
&&& \text{ } \geq 1-\eta_j, \quad \eta_j \geq 0 \\
\noalign{\quad $\forall j \in \mathcal{L_C}, \forall s_{j1}, s_{j2} \in \{-1, 1\}, s_{j1} = s_{j2} :$}
&&& \text{ } \max_{z_{j1} \neq z_{j2}} \phi(\mathbf{x}_{j1}, \mathbf{x}_{j2}, z_{j1}, z_{j2}) - \phi(\mathbf{x}_{j1}, \mathbf{x}_{j2}, s_{j1}, s_{j2})\\
&&& \text{ } \geq 1-\eta_j, \quad \eta_j \geq 0 \\
\noalign{\quad $\forall i \in \mathcal{U}:$}
&&&\text{ } \max_{y^s_i \in \{-1, 1\}} 2y^s_i\mathbf{w}^T \mathbf{x}_i \geq 1-\xi_i , \\
&&& \text{ } \xi_i \geq 0 \enspace,
\end{aligned}
\end{gathered}
\end{equation}
where $\mathcal{L_M}$ and $\mathcal{L_C}$ are the sets of ML and CL constraints, respectively, $L_\mathcal{M}$ and $L_\mathcal{C}$ are the sizes of those sets, $\eta_j$ are slack variables for the pairwise constraints, $\mathcal{U}$ is the set of unconstrained points, $U$ is the size of that set and $\xi_i$ are slack variables for the unconstrained points.  $j_1$ and $j_2$ represent the two points constrained by pairwise constraint $j$.  Here, the must-link and cannot-link constraints each impose a soft margin on the difference in score between the highest-scoring joint projection that satisfies the constraint and the highest scoring joint projection that does \emph{not} satisfy the constraint.  This formulation is sufficient for standard clustering problems, but requires some modification in order to function well in our problem setting.

\subsection{Relaxed semi-supervised max-margin clustering}
Because cannot-link constraints disappear from the hierarchy learning problem once they are satisfied (see Section~\ref{sec:HFDtrainalg}), the number of relevant cannot-link constraints will generally decrease much more quickly than the number of must-link constraints.  This will lead to highly imbalanced constraint sets in lower levels of the hierarchy. 

Under the original SSMMC formulation, imbalanced cases such as these may well yield trivial one-class solutions wherein the ML constraints are well satisfied, but the few CL constraints are highly \emph{un}satisfied.  To address this problem, we simply separate the ML and CL constraints into two distinct optimization terms, each with equal weight.

Second, and more significantly, we must modify SSMMC to handle the \emph{hierarchical} nature of our problem.  Consider a case with 4 semantic classes: apple, orange, bicycle and motorcycle.  In a binary hierarchical setting, the most reasonable way to approach this problem is to first separate apples and oranges from bicycles and motorcycles, then divide the data into pure leaf nodes lower in the tree.  

Standard SSMMC, however, will instead attempt to simultaneously satisfy the cannot-link constraints between \emph{all} of these classes, which is impossible.  As a result, the optimization algorithm may seek a compromise solution that weakly violates all or most of the constraints, rather than one that strongly satisfies a subset of the constraints and ignores the others (e.g. that separates apples and oranges from bicycles and motorcycles).

We handle this complication by relaxing the clustering algorithm to focus on only a \emph{subset} of the CL constraint set, and integrate the selection of that subset into the optimization problem.  Thus, our variant of semi-supervised MMC simultaneously optimizes $\mathbf{w}$ to satisfy a subset of the CL constraint set $\mathcal{L_C^\prime}\subset \mathcal{L_C}$, \emph{and} seeks the $\mathcal{L_C^\prime}$ that can best be satisfied by a binary linear discriminant:
\begin{equation}
\label{eq:mrmmc2}
\begin{gathered}
\shoveleft \min_{\mathbf{w}, \eta, \xi, \mathcal{L^\prime_C}} \frac{\lambda}{2}\|\mathbf{w}\|^2 + \frac{1}{L_\mathcal{M}}\sum_{j\in \mathcal{L_M}} \eta_j  + \frac{1}{ L^\prime_{\mathcal{C}}}\sum_{j\in \mathcal{L^\prime_C}} \eta_j + \frac{C}{U}\sum_{i=1}^U\xi_i \\
\begin{aligned}
&\text{s.t.}&&  \\
\noalign{\quad $\forall j \in \mathcal{L_M}, \forall s_{j1}, s_{j2} \in \{-1, 1\}, s_{j1} \neq s_{j2} :$}
&&& \text{ } \max_{z_{j1}=z_{j2}} \phi(\mathbf{x}_{j1}, \mathbf{x}_{j2}, z_{j1}, z_{j2}) - \phi(\mathbf{x}_{j1}, \mathbf{x}_{j2}, s_{j1}, s_{j2})\\
&&& \text{ } \geq 1-\eta_j, \quad \eta_j \geq 0 \\
\noalign{\quad $\exists \text{ } \mathcal{L^\prime_C} \subset \mathcal{L_C} \text{ of size } L^\prime_\mathcal{C} \text{ s.t.} :$}
\noalign{\quad \quad $\forall j \in \mathcal{L_C^\prime}, \forall s_{j1}, s_{j2} \in \{-1, 1\}, s_{j1} = s_{j2} :$}
&&& \text{ } \quad \max_{z_{j1} \neq z_{j2}} \phi(\mathbf{x}_{j1}, \mathbf{x}_{j2}, z_{j1}, z_{j2}) - \phi(\mathbf{x}_{j1}, \mathbf{x}_{j2}, s_{j1}, s_{j2})\\
&&& \text{ } \quad \geq 1-\eta_j, \quad \eta_j \geq 0 \\
\noalign{\quad $\forall i \in \mathcal{U}:$}
&&& \text{ } \max_{y^s_i \in \{-1, 1\}} 2y^s_i\mathbf{w}^T \mathbf{x}_i \geq 1-\xi_i , \\
&&& \text{ } \xi_i \geq 0 \enspace,
\end{aligned}
\end{gathered}
\end{equation}
We set the size of $\mathcal{L^\prime_C}$ via the parameter $L^\prime_\mathcal{C}$.

\subsubsection{Semi-supervised MMC optimization}

We optimize our SSMMC formulation via a Constrained Concave Convex Procedure (CCCP)~\cite{yuille2003concave, zeng2012semi}.  This is an iterative process that, at each iteration $t$, frames the constraints in (\ref{eq:mrmmc2}) as the difference of two convex functions and replaces one of those functions with its tangent at $\mathbf{w}^{(t)}$, resulting in a convex optimization problem than can be easily solved via subgradient projection.

We first use a heuristic \cite{zeng2012semi} to initialize $\mathbf{w}^{(0)}$ non-randomly based on $\mathcal{L}$.  We compute pairwise constraint scatter matrices $\mathbf{S}_\mathcal{M}$ and $\mathbf{S}_\mathcal{C}$:
\begin{align}
\mathbf{S}_\mathcal{M} &= \frac{1}{L_\mathcal{M}}\sum_{j \in \mathcal{L_M}} (\mathbf{x}_{j1} - \mathbf{x}_{j2}) (\mathbf{x}_{j1} - \mathbf{x}_{j2})^T\\
 \mathbf{S}_\mathcal{C} &= \frac{1}{L_\mathcal{C}}\sum_{j \in \mathcal{L_C}} (\mathbf{x}_{j1} - \mathbf{x}_{j2}) (\mathbf{x}_{j1} - \mathbf{x}_{j2})^T \enspace.
\end{align}
We can then use $\mathcal{S_M}$ and $\mathcal{S_C}$ to compute a projection $\mathbf{w}^{(0)}$ that attempts to maximize distance between CL and minimize distance between ML point pairs by computing the largest eigenvector of the general eigenproblem:
\begin{equation}
 \mathbf{S}_\mathcal{C}\mathbf{v} = \lambda \mathbf{S}_\mathcal{M} \mathbf{v} \enspace.
\end{equation}
Note that, because $\mathcal{L^\prime_C}$ is not available at this point in the optimization, this step must be performed on all of the cannot-link constraints.

Once $\mathbf{w}^{(0)}$ has been computed, we can begin CCCP to optimize the max-margin problem.  Given $\mathbf{w}^{(t)}$, we begin each iteration of CCCP by computing the tangents of constraint functions at $\mathbf{w}^{(t)}$:

Denote $\left(z_{j1}^{\mathcal{M}(t)}, z_{j2}^{\mathcal{M}(t)} \right)$ and $\left(z_{j1}^{\mathcal{C}(t)}, z_{j2}^{\mathcal{C}(t)} \right)$ as the best cluster assignments under $\mathbf{w}^{(t)}$ that satisfy their associated constraint $j$:
\begin{align}
\left(z_{j1}^{\mathcal{M}(t)}, z_{j2}^{\mathcal{M}(t)} \right) &= \argmax_{z_{j1} = z_{j2} | j \in \mathcal{L_M}} \phi_t(\mathbf{x}_{j1}, \mathbf{x}_{j2}, z_{j1}, z_{j2}) \label{eq:bestml}\\
\left(z_{j1}^{\mathcal{C}^\prime(t)}, z_{j2}^{\mathcal{C}^\prime(t)} \right) &= \argmax_{z_{j1} \neq z_{j2} | j \in \mathcal{L}_\mathcal{C}^\prime} \phi_t(\mathbf{x}_{j1}, \mathbf{x}_{j2}, z_{j1}, z_{j2}) \label{eq:bestcl}
\end{align}
Similarly, denote $y_i^{(t)}$ as the best unary cluster assignment label for $\mathbf{x}_i$ under $\mathbf{w}^{(t)}$:
\begin{equation}
\label{eq:ssmmcyt}
y_i^{(t)} = \argmax_{y_i \in \{-1, 1\}} y_i \mathbf{w}^{(t)T}\mathbf{x}_i
\end{equation}
Finally, we select a candidate $\mathcal{L}^{\prime(t)}_{\mathcal{C}}$ by choosing the $ L^\prime_{\mathcal{C}}$ cannot-link constraints with the largest satisfaction margin (again, under $\mathbf{w}^{(t)}$):
\begin{equation}
\begin{gathered}
\label{eq:lcprime}
\begin{aligned}
\mathcal{L}^{\prime(t)}_{\mathcal{C}} = \argmax_{\mathcal{L}^{\prime}_{\mathcal{C}} \subset \mathcal{L_C}, \left|\mathcal{L}^{\prime}_{\mathcal{C}}\right| = L^\prime_{\mathcal{C}}}
\text{ } \sum_{j \in \mathcal{L}^{\prime}_{\mathcal{C}}}
\min_{s_{j1}=s_{j2}} & \Big[ \phi_t\left(\mathbf{x}_{j1}, \mathbf{x}_{j2}, z_{j1}^{\mathcal{C}(t)}, z_{j2}^{\mathcal{C}(t)}\right) \\
& - \phi_t(\mathbf{x}_{j1}, \mathbf{x}_{j2}, s_{j1}, s_{j2}) \Big]
\end{aligned}
\end{gathered}
\end{equation}

By setting all of these values as constants, we obtain a convex (but non-differentiable) optimization problem:
\begin{equation}
\label{eq:SSMMC}
\begin{gathered}
\shoveleft \min_{\mathbf{w}^{(t)}, \eta, \xi} \frac{\lambda}{2}  \|\mathbf{w}^{(t)}\|^2 + \frac{1}{L_\mathcal{M}}\sum_{j\in \mathcal{L_M}} \eta_j  + \frac{1}{ L^\prime_{\mathcal{C}}}\sum_{j\in \mathcal{L}^{\prime(t)}_{\mathcal{C}}} \eta_j + \frac{C}{U}\sum_{i=1}^U\xi_i \\
\begin{aligned}
&\text{s.t.}&&  \\
\noalign{\quad $\forall j \in \mathcal{L_M}, \forall s_{j1}, s_{j2} \in \{-1, 1\}, s_{j1} \neq s_{j2} :$}
&&& \text{ } \phi_t(\mathbf{x}_{j1}, \mathbf{x}_{j2}, z_{j1}^{\mathcal{M}(t)}, z_{j2}^{\mathcal{M}(t)}) - \phi_t(\mathbf{x}_{j1}, \mathbf{x}_{j2}, s_{j1}, s_{j2})\\
&&& \text{ } \geq 1-\eta_j, \quad \eta_j \geq 0 \\
\noalign{\quad $\forall j \in  \mathcal{L}^{\prime(t)}_{\mathcal{C}}, \forall s_{j1}, s_{j2} \in \{-1, 1\}, s_{j1} = s_{j2} :$}
&&& \text{ } \phi_t(\mathbf{x}_{j1}, \mathbf{x}_{j2}, z_{j1}^{\mathcal{C}(t)}, z_{j2}^{\mathcal{C}(t)}) - \phi_t(\mathbf{x}_{j1}, \mathbf{x}_{j2}, s_{j1}, s_{j2})\\
&&& \text{ } \geq 1-\eta_j, \quad \eta_j \geq 0 \\
\noalign{\quad $\forall i \in \mathcal{U}:$}
&&& \text{ } 2y^{(t)}_i\mathbf{w}^{(t)T} \mathbf{x}_i \geq 1-\xi_i , \\
&&& \text{ } \xi_i \geq 0 \enspace,
\end{aligned}
\end{gathered}
\end{equation}

This problem can be efficiently solved via subgradient projection.  At each subgradient iteration $r$, $\nabla_r$ can be computed via:
\begin{equation}
\begin{gathered}
\begin{aligned}
\label{eq:nabla_semisup}
\nabla_r = \text{ }& \lambda \mathbf{w} \\
&+ \frac{1}{L_\mathcal{M}} \sum_{j \in \mathcal{L}_\mathcal{M}^{(r)}}  \Big[ \left( s_{j1}^{\mathcal{M}^r}x_{j1} +  s_{j2}^{\mathcal{M}^r}x_{j2} \right) \\
& - \left( z_{j1}^{\mathcal{M}(t)}x_{j1} +  z_{j2}^{\mathcal{M}(t)}x_{j2} \right) \Big]   \\
&+ \frac{1}{L_\mathcal{C}^\prime} \sum_{j \in \mathcal{L}_\mathcal{C}^{\prime(r)}}  \Big[ \left( s_{j1}^{\mathcal{C}^{\prime r}}x_{j1} +  s_{j2}^{\mathcal{C}^{\prime r}}x_{j2} \right) \\
& - \left( z_{j1}^{\mathcal{C}^\prime(t)}x_{j1} +  z_{j2}^{\mathcal{C}^\prime(t)}x_{j2} \right) \Big]   \\
&+ \frac{C}{U}\sum_{i \in \mathcal{U}} y_i^{(t)}\mathbf{x}_i \mathbbm{1}\left( 2y^{(t)}_i\mathbf{w}^T\mathbf{x}_i \leq 1\right)  \enspace,
\end{aligned}
\end{gathered}
\end{equation}
where $\mathcal{L}_\mathcal{M}^{(r)}$ and $\mathcal{L}_\mathcal{C}^{\prime(r)}$ are the set of pairwise constraints with non-zero $\eta_j$ values under $\mathbf{w}^{(r)}$:
\begin{align}
\mathcal{L}_\mathcal{M}^{(r)} = &\Big\{j \in \mathcal{L_M} \Big| \phi_r\left(\mathbf{x}_{j1}, \mathbf{x}_{j2}, z_{j1}^{\mathcal{M}(t)}, z_{j2}^{\mathcal{M}(t)}\right) \nonumber \\
 & - \max_{s_{j1} \neq s_{j2}} \phi_r\left(\mathbf{x}_{j1}, \mathbf{x}_{j2}, s_{j1}, s_{j2}\right)   < 1 \Big\} \\
\mathcal{L}_\mathcal{C}^{\prime(r)} = &\Big\{j \in \mathcal{L}_\mathcal{C}^\prime \Big| \phi_r\left(\mathbf{x}_{j1}, \mathbf{x}_{j2}, z_{j1}^{\mathcal{C}^\prime(t)}, z_{j2}^{\mathcal{C}^\prime(t)}\right) \nonumber \\
 & - \max_{s_{j1} \neq s_{j2}} \phi_r\left(\mathbf{x}_{j1}, \mathbf{x}_{j2}, s_{j1}, s_{j2}\right)   < 1 \Big\} \enspace,
\end{align}
 and we define the highest-scoring constraint-\emph{violating} assignments for each pair:
\begin{align}
\left( s_{j1}^{\mathcal{M}^r}, s_{j2}^{\mathcal{M}^r} \right) &= \argmax_{j \in \mathcal{L_M}, s_{j1} \neq s_{j2}} \phi_r\left(\mathbf{x}_{j1}, \mathbf{x}_{j2}, s_{j1}, s_{j2}\right) \\
\left( s_{j1}^{\mathcal{C}^{\prime r}}, s_{j2}^{\mathcal{C}^{\prime r}} \right) &= \argmax_{j \in \mathcal{L}_\mathcal{C}^\prime, s_{j1} = s_{j2}} \phi_r\left(\mathbf{x}_{j1}, \mathbf{x}_{j2}, s_{j1}, s_{j2}\right) \enspace.
\end{align}
Finally, it is shown in~\cite{zeng2012semi} that the optimal solution to $\mathbf{w}$ is bounded by:
\begin{equation}
\mathbf{w}^* \in \left\{ \mathbf{w} \Bigg| \|\mathbf{w}\| \leq \rho =\sqrt{\frac{1+C}{\lambda}} \right\} \enspace,
\end{equation}
so we project $\mathbf{w}^{(r)}$ back into this space at each iteration.

\begin{algorithm}
\caption{Subgradient optimization of $\mathbf{w}$ in semi-supervised MMC \label{alg:SSMMCsubgrad}}
\begin{algorithmic}
\State $r \gets 0$
\State randomly initialize $\mathbf{w}^{(r)}$, s.t. $\|\mathbf{w}^{(r)}\| \leq \rho$
\While{$\mathbf{w}^{(r)}$ not converged}
	\State Compute $\nabla_r$ via (\ref{eq:nabla_semisup})
	\State $\mathbf{w}^{(r+\frac{1}{2})} \gets \mathbf{w}^{(r)} + \frac{1}{\lambda r}\nabla_r$ 
	\State $\mathbf{w}^{(r+1)} \gets \min \left(1, \rho / \left\| \mathbf{w}^{(r+\frac{1}{2})}\right\| \right) \mathbf{w}^{(r+\frac{1}{2})}$
	\If{$\|\mathbf{w}^{(r+1)} - \mathbf{w}^{(r)}\| / \max (\|\mathbf{w}^{(r)}\|,  \|\mathbf{w}^{(r+1)}\| ) \leq \epsilon_1$}
		\State $\mathbf{w}^{(r)}$ is converged
	\EndIf
	\State $r \gets r + 1$
\EndWhile
\end{algorithmic}
\end{algorithm}

The subgradient optimization method for solving (\ref{eq:SSMMC}) is described in Algorithm \ref{alg:SSMMCsubgrad} and the full CCCP procedure for semi-supervised MMC is described in Algorithm \ref{alg:SSMMC} (note that we set $C=0$ for the first three iterations in order to allow the more reliable supervised constraints to guide the optimization to a strong solution region, before introducing the unsupervised constraints to refine it).

\begin{algorithm}
\caption{CCCP optimization for semi-supervised MMC \label{alg:SSMMC}}
\begin{algorithmic}
\State $t \gets 0$
\State randomly initialize $\mathbf{w}^{(t)}$, s.t. $\|\mathbf{w}^{(r)}\| \leq \rho$
\While{$\mathbf{w}^{(t)}$ not converged}
	\If{t < 3}
		\State $C^{(t)} \gets 0$
	\Else
		\State $C^{(t)} \gets C$
	\EndIf
	\State Compute $y^{(t)}$ via (\ref{eq:ssmmcyt})
	\State Compute $\left(z_{j1}^{\mathcal{M}(t)}, z_{j2}^{\mathcal{M}(t)} \right)$ via (\ref{eq:bestml})
	\State Compute $\left(z_{j1}^{\mathcal{C}(t)}, z_{j2}^{\mathcal{C}(t)} \right)$ via (\ref{eq:bestcl})
	\State Compute $\mathcal{L}^{\prime(t)}_{\mathcal{C}}$ via (\ref{eq:lcprime})
	\State Compute $w^{(t+1)}$ via Algorithm \ref{alg:SSMMCsubgrad}
\If{$\|\mathbf{w}^{(t+1)} - \mathbf{w}^{(t)}\| / \max (\|\mathbf{w}^{(t)}\|,  \|\mathbf{w}^{(t+1)}\| ) \leq \epsilon_2$}
\State $\mathbf{w}^{(t)}$ is converged
\EndIf
\State $t \gets t + 1$
\EndWhile
\end{algorithmic}
\end{algorithm}

\subsection{Unsupervised MMC}
Because HFD progressively eliminates constraints as the hierarchy grows deeper (see Section~\ref{sec:HFDtrainalg}), at lower levels of the hierarchy we may encounter nodes where there are no cannot-link constraints available, and hence SSMMC cannot proceed~\cite{zeng2012semi}.  In these cases, we fall back on the unsupervised Membership Requirement MMC (MRMMC) formulation proposed by Hoai and De la Torre~\cite{hoai2012maximum}.  We optimize this problem via block-coordinate descent as described in that work.

\section{Fast approximate nearest neighbors in hierarchy metric space}
\label{sec:hfdnn}

One problem with this approach is the potentially high (though still embarrassingly parallel) cost of computing each pairwise distance, as compared to a Euclidean or even Mahalanobis distance.  This is worsened, for many applications, by the unavailability of traditional fast approximate nearest-neighbor methods (e.g. kd-trees~\cite{bentley1975multi}, hierarchical $k$-means~\cite{muja2009fast} or hashing~\cite{gionis1999similarity}), which require an explicit representation of the data in the metric space in order to function.

We address the latter problem by introducing our own fast approximate nearest-neighbor algorithm, which takes advantage of the tree-based structure of the metric to greatly reduce the number of pairwise distance computations needed to compute a set of nearest-neighbors for a query point $x$.

We begin by tracing the path taken by $x$ through each tree in the forest, and thus identifying each leaf node containing $x$.  We then seek $k_\mathcal{O}$ candidate neighbors from each tree, beginning by sampling other training points from the identified leaf nodes, then, if necessary, moving up the tree parent-node-by-parent-node until $k_ \mathcal{O}$ candidates have been found.  The candidate sets from each tree are then combined to yield a final candidate neighbor set  $\mathcal{O}$, such that  $|\mathcal{O}| \leq T\cdot k_\mathcal{O}$.  We then compute the full hierarchy distance $D(x, y)$ for all $y \in  \mathcal{O}$, sort the resulting distances, and return the $k$ closest points.

This approximation method functions by assuming that, intuitively, a point's nearest-neighbors within the full forest metric space are very likely to \emph{also} be nearest-neighbors under at least \emph{one} of the individual tree metrics.  We evaluate this method empirically on several small-to-midsize datasets, and the results strongly support the validity of this approximation (Section~\ref{sec:expnn}).

\section{Complexity Analysis}
\label{sec:complex}
\textbf{HFD training}

The overall complexity of semi-supervised max-margin clustering is $\mathbf{O}(d^3 + nd)$~\cite{zeng2012semi} (where $n$ is the total number of constraints plus the number of unconstrained points).  If we ignore $d$ (which in our case is replaced by the parameter $d_k$, and generally speaking $d_k \ll d$), this leaves SSMMC as a $\mathbf{O}(n)$ operation.  In the tree setting, we are using SSMMC in a divide-and-conquer fashion---thus, if we assume that we divide the data roughly in half with each SSMMC operation, the complexity of fully training an HFD tree is $\mathbf{O}(n\log n)$, and the total training cost is thus $\mathbf{O}(Tn\log n)$.  Given the embarrassingly parallel nature of the problem, the $T$ factor can be ignored in many cases, allowing an HFD model to be trained in $\mathbf{O}(n\log n)$ time.

\textbf{HFD inference}

Computing a single HFD metric distance requires traversing down each tree in the forest one time, for an (again embarrassingly parallel) complexity cost of $\mathbf{O}(T\log n)$.  Many of the most common applications of a metric require computing nearest-neighbors between the training set and a test set of size $m$.  This requires $mn$ distance evaluations, so a brute force nearest-neighbor search under HFD costs $\mathbf{O}(mnT\log n)$, or $\mathbf{O}(nT\log n)$ for a single test point.

Our approximate nearest-neighbor algorithm significantly reduces this cost.  Computing candidate neighbors for a single point costs only $\mathbf{O}(T\log n)$.  There will be at most $Tk_\mathcal{O}$ candidates for each set, so the cost for computing distances to each candidate is $\mathbf{O}(T^2k_\mathcal{O}\log n)$, or $\mathbf{O}(T^2\log n)$ if we ignore the parameter.  It should be noted, though, that in practice there is significant overlap between the candidate sets returned by different trees, and this overlap increases with $T$, so the actual cost of this step is generally much lower.  

Thus, the complexity of our approximate nearest-neighbor method, when applied to an entire dataset, is $\mathbf{O}(mT^2\log n)$.  Even in the worst case, this is an improvement (since $T \ll n$ on even moderately sized datasets), and in practice is generally much better than the worst case.

\section{Experiments}
\label{sec:exp}

Below we present several experiments quantifying HFD's performance.  First, we validate the accuracy and efficiency of our approximate nearest-neighbor retrieval method.  We then carry out benchmark comparisons against other state-of-the-art metric learning techniques in the $k$-nearest neighbor classification, large-scale image retrieval and semi-supervised clustering domains.

\subsection{Datasets}

We use a range of datasets, from small- to large-scale, to evaluate our method.  For small to mid-range data, we use a number of well-known UCI sets~\cite{Bache+Lichman:2013}: sonar (208 x 60 x 2), ionosphere (351 x 34 x 2), balance (625 x 4 x 3), segmentation (2,310 x 18 x 7) and magic (19,020 x 10 x 2), as well as the USPS handwritten digits dataset (11,000 x 256 x 10)~\cite{hull1994database}.

For our larger scale experiments, we relied on the CIFAR tiny image datasets~\cite{krizhevsky2009learning}. CIFAR-10 consists of 50,000 training and 10,000 test images, spread among 10 classes.  CIFAR-100 also contains 50,000 training and 10,000 testing images, but has 2 different label sets---a coarse set with 20 classes, and a fine set with 100 classes.  All CIFAR instances are 36x36 color images, which we have reduced to 300 features via PCA.

In all our experiments, the data is normalized to 0 mean and unit variance before any metric methods are applied to it.

\subsection{Approximate nearest-neighbor retrieval}
\label{sec:expnn}

Because we use it for retrieval in all of our other experiments, we first evaluate the accuracy cost and efficiency benefits of our approximate nearest-neighbor method.  We evaluate accuracy by training an HFM  model with 100 trees.  We then return 50 approximate nearest-neighbors for each point in the dataset and compute mean average precision (mAP) relative to the ground truth 50 nearest-neighbors (obtained via brute force search).  Average precision scores are computed at 10, 20, 30, 40 and 50 nearest-neighbors.  We do retrieval at $k_\mathcal{O}=$ 1, 3, 5, 10, 20 and 30, and report both the mAP results and the time taken (as a proportion of the brute force time) at each value on several datasets.

\begin{table}[htbp]
\centering
\caption{Approximate nearest-neighbor retrieval mAP scores}
\begin{tabular}{|l|c|c|c|}
\hline
 $k_\mathcal{O}$ & \textbf{Sonar} & \textbf{Seg.} & \textbf{USPS} \\ \hline
1 & 0.812 & 0.715 & 0.6559 \\ 
3 & 0.965 & 0.904 & 0.8700 \\ 
5 & 0.987 & 0.956 & 0.9274 \\ 
10 & 0.997 & 0.987 & 0.9710 \\ 
20 & 0.998 & 0.994 & 0.9897 \\ 
30 & 0.998 & 0.995 & 0.9945 \\ \hline
\end{tabular}
\end{table}

\begin{table}[htbp]
\centering
\caption{Approximate nearest-neighbor retrieval times (as a proportion of brute force search time)}
\begin{tabular}{|l|c|c|c|}
\hline
 $k_\mathcal{O}$ & \textbf{Sonar} & \textbf{Seg.} & \textbf{USPS} \\ \hline
1 & 0.499 & 0.042 & 0.014 \\ 
3 & 0.547 & 0.074 & 0.034 \\ 
5 & 0.729 & 0.097 & 0.049 \\ 
10 & 0.860 & 0.147 & 0.076 \\
20 & 1.002 & 0.221 & 0.112 \\ 
30 & 0.997 & 0.270 & 0.136 \\ \hline
\end{tabular}
\end{table}

The results clearly show that our approximation method can yield significant reductions in retrieval time on larger datasets, and does so with minimal loss of accuracy.  Note that all other results we report for HFD are generated using this approximation method.  We use $k_\mathcal{O}=5$ for the CIFAR datasets, and $k_\mathcal{O}=10$ for all other data.

\subsection{Comparison methods and parameters}

\begin{figure*}[htb]
\centering
\includegraphics[width=0.8\linewidth]{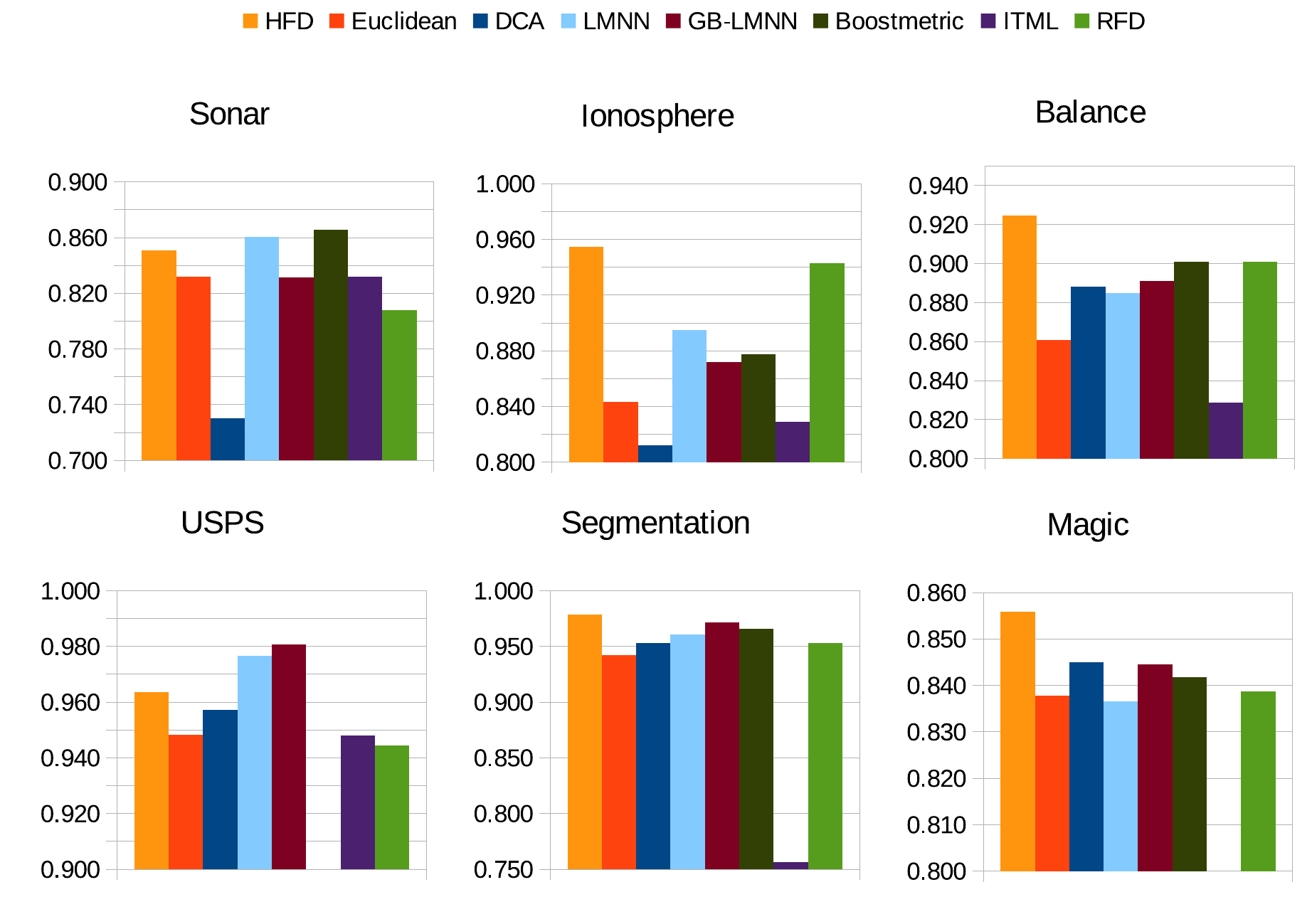}
\caption{5-nearest neighbor classification accuracy under HFD and benchmark metrics. HFD achieves the highest accuracy among tested methods on 4 out of 6 datasets, and is competitive on the remaining two. (View in color)}
\label{fig_class}
\end{figure*}

\begin{figure*}[htb]
\centering
\includegraphics[width=0.8\linewidth]{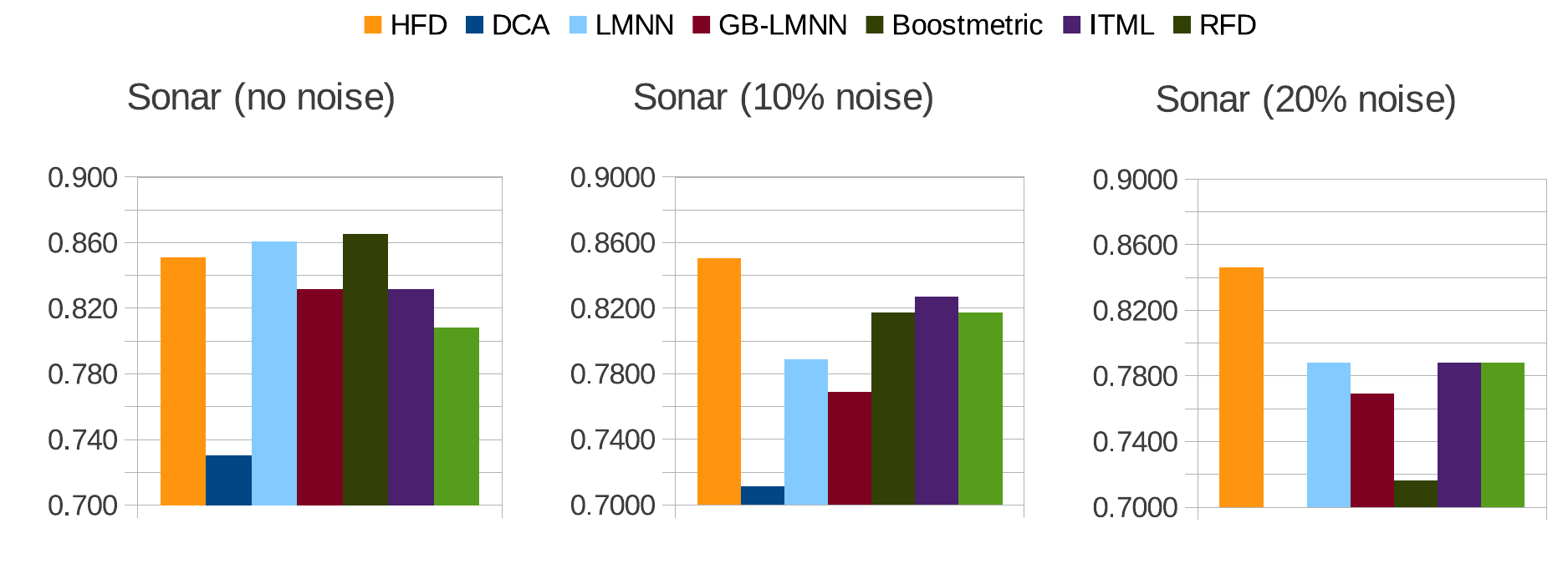}
\caption{5-nearest neighbor classification accuracy on the sonar dataset with varying levels of training label noise. While HFD does not achieve the best result in the noiseless case, it is far more robust to label noise than any other method tested. (View in color)}
\label{fig_noise}
\end{figure*}

In the following experiments, we compare our HFD model against a number of state-of-the-art metric learning techniques:  DCA~\cite{hoi2006learning}, LMNN~\cite{blitzer2005distance, shen2010scalable}, GB-LMNN \cite{kedem2012non}, ITML~\cite{davis2007information}, Boostmetric~\cite{Shen2009PSD} and RFD~\cite{xiong2012random}.  With the exception of RFD and GB-LMNN (both of which incorporate tree structures into their metrics), all are Mahalanobis methods that learn purely linear transforms of the original data.

For all ITML experiments, we cross-validated across 5 different $\gamma$ values and reported the best results.

We did not extensively tune any of the hyperparameters for HFD, instead using a common set of values (or rules for assigning values) for all datasets.  We set $T=500$ (for HFD, RFD and GB-LMNN), $d_k=\frac{d}{3}$ (with the exception of the balance dataset, where we use $d_k=d$), $L^\prime_\mathcal{C}=0.25 L_\mathcal{C}$, $\lambda=0.01$, $C=1$, $\epsilon_1=\epsilon_2=0.01$ and $\alpha=0.5$.  As a stop criteria for tree training, we set a minimum node size of 5 for the clustering and classification experiments, and 30 for the retrieval experiments.

\subsection{Nearest neighbor classification}
\label{sec:exp_class}
We next test our method using $k$-nearest neighbor classification (we use $k=5$ for all datasets).  Each dataset is evaluated using 5-fold cross-validation.  For the weakly-supervised methods, in each fold we use 1,000 constraints per class (drawn from the training data) for the sonar, ionosphere, balance and segmentation datasets.  For USPS and magic, we use 30,000 constraints in each fold.  Our results are shown in Figure \ref{fig_class}.

We found that HFD achieved the best score on 4 out of the 6 datasets tested, and was competitive on the remaining two (sonar and USPS).

We also performed some additional experiments to test the robustness of our approach compared to other methods.  We tested 5-nearest neighbor classification on the sonar dataset with varying amounts of noise added to the training labels (for consistency, we used the same noise data for all metrics).  Our results are shown in Figure \ref{fig_noise}.

While our method is only minimally effected by the added noise, the performance of all other metrics drops dramatically.  Though HFD does not obtain the best results in the noiseless case, with just 10\% of the training labels corrupted our results become significantly better than all other metrics.  This effect is even more pronounced at 20\% noise.

\subsection{Retrieval}
\begin{figure*}[htb]
\centering
\includegraphics[width=0.8\linewidth]{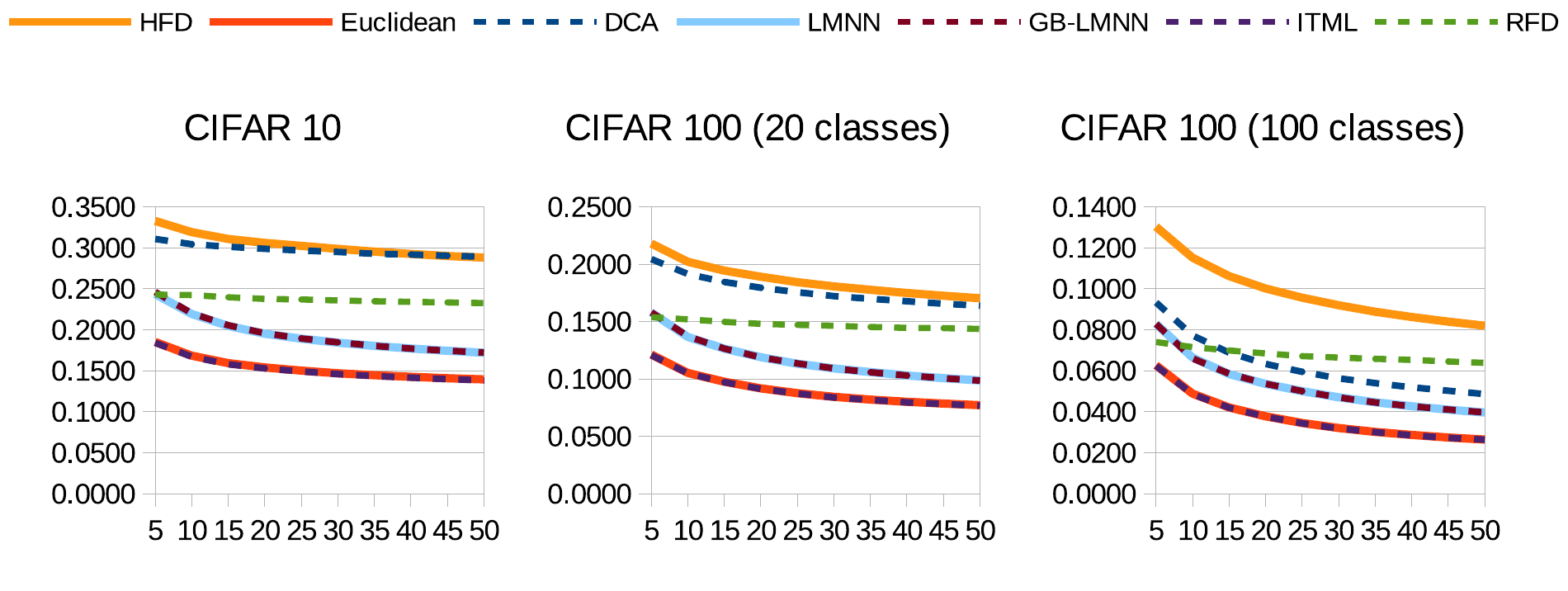}
\caption{Large-scale semantic image retrieval results for our method and benchmarks.  Only DCA is competitive with our method on the 10 and 20 class datasets, and HFD significantly outperforms all other algorithms on the 100 class problem. (View in color)}
\label{fig_retrieval}
\end{figure*}

To evaluate our method's performance (as well as, implicitly, the effectiveness of our approximate nearest-neighbor algorithm) on large-scale tasks, we computed semantic retrieval precision on labeled CIFAR tiny image datasets.  For the weakly-supervised methods, we sample 600,000 cons\-traints from the training data (which is less than 0.1\% of the full constraint set).  We do not report Boostmetric results on these sets because we were unable to obtain them.

Our results can be found in Figure \ref{fig_retrieval}, which shows retrieval accuracy at 5 through 50 images retrieved on each dataset.  HFD is clearly the best-performing method across all 3 problems.  While DCA is competitive with HFD on the 10-class and 20-class sets, this performance drops off significantly on the more difficult 100-class problem.

The particularly strong performance of HFD on the 100-class problem may be due to the relaxed SSMMC formulation, which allows our method to effectively divide the very difficult 100-class discrimination problem into a sequence of many broader, easier problems, and thus make more effective use of its cannot-link constraints than the other metrics.

\subsection{Semi-supervised clustering}

\begin{table*}[htbp]
  \centering
  \caption{Semi-supervised clustering results (V-Measure)}
\begin{tabular}{|l|ccc|ccc|}
\hline
& \multicolumn{3}{c|}{\textbf{Sonar}} & \multicolumn{3}{c|}{\textbf{Balance}} \\ 
 & \textbf{60} & \textbf{120} & \textbf{180} & \textbf{45} & \textbf{90} & \textbf{180} \\ \hline
Euclidean & 0.0493 & 0.0493 & 0.0493 & 0.2193 & 0.2193 & 0.2193 \\ \hline
DCA & 0.0959 & 0.1098 & 0.1386 & 0.0490 & 0.2430 & 0.3817 \\ \hline
ITML & 0.0650 & 0.0555 & 0.0644 & 0.2221 & 0.1915 & 0.2155 \\ \hline
RFD & 0.0932 & 0.1724 & 0.2699 & 0.1398 & 0.1980 & 0.3004 \\ \hline
HFD & \textbf{0.1267} & \textbf{0.2296} & \textbf{0.3518} & \textbf{0.3059} & \textbf{0.5128} & \textbf{0.6149} \\ \hline
\multicolumn{7}{l}{}\\ \hline
& \multicolumn{3}{c|}{\textbf{Segmentation}} & \multicolumn{3}{c|}{\textbf{USPS}} \\ 
 & \textbf{70} & \textbf{175} & \textbf{350} & \textbf{3k} & \textbf{5k} & \textbf{10k} \\ \hline
Euclidean & 0.6393 & 0.6393 & 0.6393 & 0.6493 & 0.6493 & 0.6493 \\ \hline
DCA & 0.0510 & 0.2537 & 0.6876 & 0.5413 & 0.4359 & 0.4473 \\ \hline
ITML & 0.5682 & 0.6365 & 0.5931 & 0.6447 & 0.6445 & 0.6420 \\ \hline
RFD & \textbf{0.7887} & \textbf{0.8157} & \textbf{0.8367} & \textbf{0.8248} & \textbf{0.8402} & 0.8745 \\ \hline
HFD & \textbf{0.7788} & \textbf{0.8090} & \textbf{0.8367} & 0.7258 & 0.7397 & \textbf{0.9087} \\ \hline
\end{tabular}
  \label{tab:clustsmallvmeasure}%
\end{table*}%

In order to analyze the metrics holistically, in a way that takes into account not just ordered rankings of distances but the relative values of the distances themselves, we began by performing semi-supervised clustering experiments. We sampled varying numbers of constraints from each of the datasets presented and used these constraints to train the metrics.  Note that only weakly- or semi-supervised metrics could be evaluated in this way, so only DCA, ITML, RFD and HFD were used in this experiment.

After training, the learned metrics were applied to the dataset and used to retrieve the 50 nearest-neighbors and corresponding distances for each point.  RFD and HFD return distances on a 0-1 scale, so we converted those to similarities by simply subtracting from 1.  For the other methods, the distances were converted to similarities by applying a Gaussian kernel (we used $\sigma=$ 0.1, 1, 10, 100 or 1000---whichever yielded the best results for that metric and dataset).  

We then used the neighbor and similarity data to construct a number of sparse similarity matrices from varying numbers of nearest-neighbors (ranging from 5 to 50) and computed a spectral clustering~\cite{shi2000normalized} solution for each.  We evaluated these clustering outputs using V-Measure~\cite{rosenberg2007v} and recorded the best result for each metric-dataset pair (see Table \ref{tab:clustsmallvmeasure}---the numbers below the dataset names indicate the number of constraints used in that test).

The tree based methods, RFD and HFD, demonstrated a consistent and significant advantage on this data.  Between the two tree-based methods, HFD yielded better results on the sonar and balance data, while both were competitive on the segmentation and USPS datasets.

It is notable that the difference between the euclidean performance and that of the tree-based metrics is much more pronounced in the clustering domain.  This would suggest that the actual distance values (as opposed to the distance rankings) returned by the tree-based metrics contain much stronger semantic information than those returned by the linear methods.

\section{Conclusion}
In this paper, we have presented a novel semi-supervised nonlinear distance metric learning procedure based on forests of cluster hierarchies constructed via an iterative max-margin clustering procedure with a novel relaxed constraint formulation.  Our experimental results show that that this algorithm is competitive with the state-of-the-art on small- and medium-scale datasets, and potentially superior for large-scale problems.  We also present a novel in-metric approximate nearest-neighbor retrieval algorithm for our method that greatly decreases retrieval times for large data with little reduction in accuracy.

In the future, we hope to expand this metric to less-well-explored learning settings, such as those with more complex semantic relationship structures (e.g., hierarchies or ``soft'' class membership).   By extending our method to incorporate relative similarity triplet constraints, we could allow semi-supervised metric learning even in these domains where binary pairwise constraints are no longer possible.

{
\bibliographystyle{unsrt}
\bibliography{arXiv_JoXiCo_hiermetric}
}

\end{document}